\documentclass{article}
\usepackage{ijcai20}

\usepackage{times}

\usepackage{soul}
\usepackage{url}
\usepackage[hidelinks]{hyperref}
\usepackage[utf8]{inputenc}
\usepackage[small]{caption}
\usepackage{graphicx}
\usepackage{amsmath}
\usepackage{booktabs}
\usepackage{amssymb}

\title{Machine learning for improving performance in an evolutionary algorithm for minimum path with uncertain costs given by massively simulated scenarios}

\author{
Ricardo Di Pasquale$^1$\and
Javier Marenco$^2$\\
\affiliations
$^1$Facultad de Ingenier\'ia y Ciencias Agrarias, Pontificia Universidad Cat\'olica Argentina, Argentina.\\
$^2$Instituto de Ciencias, Universidad Nacional de General Sarmiento, Argentina.\\
\emails
rdipasquale@uca.edu.ar,
jmarenco@ungs.edu.ar
}

\begin{document}

\maketitle

\begin{abstract}
In this work we introduce an implementation for which machine learning techniques helped improve the overall performance of an evolutionary algorithm for an optimization problem, namely a variation of robust minimum-cost path in graphs. In this \emph{big data optimization} problem, a path achieving a good cost in most scenarios from an available set of scenarios  --generated by a simulation process-- must be obtained.

The most expensive task of our evolutionary algorithm, in terms of computational resources, is the evaluation of candidate paths: the fitness function must calculate the cost of the candidate path in every generated scenario.  Given the large number of scenarios, this task must be implemented in a distributed environment.

We implemented gradient boosting decision trees to classify candidate paths in order to identify good candidates. The cost of the not-so-good candidates is simply forecasted. We studied the training process, gain performance, accuracy, and other variables. Our computational experiments show that the computational performance was significantly improved at the expense of a limited loss of accuracy. 
\end{abstract}

\section{Introduction}

In this work we report a case in which machine learning (ML) can successfully boost the performance (in terms of running time) of a metaheuristic search in a big data environment. We are particularly focused on instances of \emph{big data optimization} (BDO) problems, namely optimization problems where massive or complex data arises as an insurmountable problem for traditional approaches. 

In this work we deal with an evolutionary algorithm searching for minimum-cost paths in a large number of graphs, as many as generated scenarios, simultaneously. The use of ML has resulted in a reduction of computing resources, as well as an improvement in the algorithm performance in terms of overall running time. The remainder of the paper is organized as follows. The studied problem will be stated in Section 2. In Section 3 we introduce the scenario-generating procedure. Section 4 describes our evolutionary algorithm, whereas the evolutionary workflow implementation details are described in Section 5. In Section 6 we analyze considerations coming from problem sizing. Results are presented in Section 7. Finally, in Section 8, machine learning algorithms are presented in order to search for performance improvements. Conclusions are described in Section 9.

\section{The problem}

We consider in this work the \emph{sailing regatta route optimization problem} introduced in \cite{1}. We are given an acyclic directed graph $G = (V,E)$ representing possible paths between the start and goal points in the regatta. The vertex set $V$ is defined by a bidimensional discretization of the regatta's geographic map (called the \emph{court} in this context). The edge set $E$ represents possible navigation maneuvers between neighboring vertices (including keeping course with no maneuvers). The costs associated with the edges model the expected navigation times including maneuvering costs.


The graph $G$ is based on the court's map. The court is divided into fixed square cells (in our case, we use 50 m${}^2$ cells). Fig.~1 shows the basic graph, where navigable cells are colored in blue, and non-navigable cells are colored in pink. Fig.~2 shows possible state transitions (i.e., edges) within one cell.

\begin{figure}[!htb]
\centering
\includegraphics[scale=0.5]{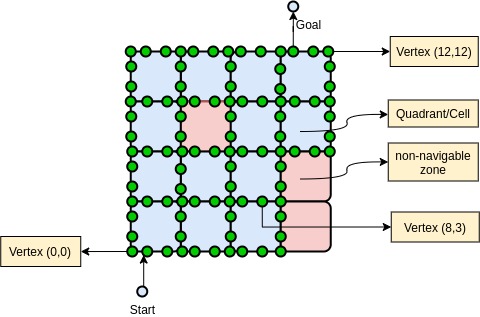}
\caption{Basic graph model for a simple court.}
\end{figure}

\begin{figure}[!htb]
\centering
\includegraphics[scale=0.3]{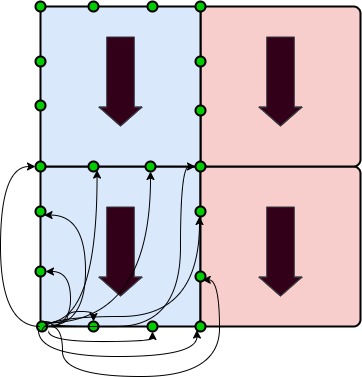}
\caption{Possible state transitions within one cell.}
\end{figure}

Sailboat maneuvers include \textit{tack, gybe, bow-up} and \textit{bow-down}. Not all maneuvers are possible all the time, as available maneuvers depend on the wind angle. In \cite{1}, this simple model is extended in order to take into account possible navigation maneuvers. This is accomplished by adding a new dimension to the basic graph model of Fig.~1, which represents the \emph{sailing mode} so only edges with available maneuvers are included in the obtained graph. Fig.~3 shows the three-dimensional graph for a simple example, representing the geographic map as well as all possible maneuvers. It is important to notice that surface ocean current effects on the sailboats are considered as not important in this approach, given that they may affect all regatta sailboats in the same way within short periods of time.

\begin{figure}[!htb]
\centering
\includegraphics[scale=0.3]{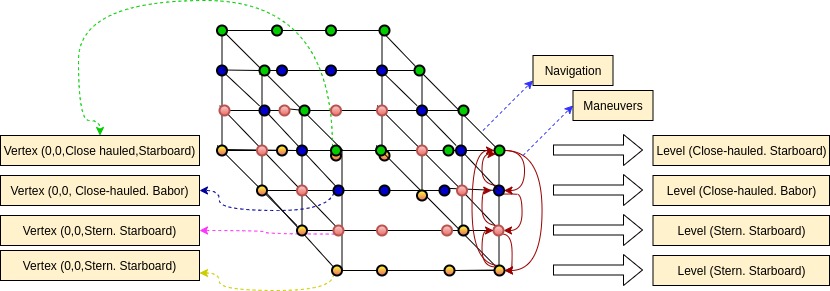}
\caption{Complete court graph representation.}
\end{figure}

Every edge $e\in E$ has an associated cost $C_e = NC_e+MC_e$ where $NC_e$ is the navigation cost and $MC_e$ is the maneuver cost for the edge $e$, respectively. The details of costs calculations are out of scope for this paper, and we refer to \cite{1} for details.

Within this context and a known \textit{a priori} static wind scenario $W_0$, the authors in \cite{1} provide two approaches: (1) an exact algorithm to find minimum-cost routes (i.e., an optimal solution $s_{w0}$ for $W_0$) and (2) a real-time heuristic using sailboat navigation tools as input.

\section{Simulated wind scenarios}

The accuracy of approach (1) in \cite{1} depends on (a) the regatta duration, (b) the court area, and (c) weather stability. 
Light fluctuations of these factors during the regatta can render an optimal solution $s_{w0}$ obsolete. 
Furthermore, in a time-dependent approach, the bigger the fluctuations of these factors, the worse $s_{w0}$ can finally be. Even if the real-time  heuristic (2) is used to complement the optimal solution $s_{w0}$, it is impossible to discard a decision after it has been taken, i.e., no rollback for maneuvers exists in this context.

In consequence, we propose in this work a scenario-based approach based on the observation that winds in soil, river, and seas have a higher variation rate than winds in higher atmosphere layers. We locate the court in a specific location in order to provide accurate models, and we have chosen the ``Rio de la Plata'' estuary near Buenos Aires, Argentina (specially due to the fact that we can easily access wind observations within this area). Fig.~4 shows our proposed location for a 6.25 km${}^2$ court ($50 \times 50$ cells).

\begin{figure}[!htb]
\centering
\includegraphics[scale=1.15]{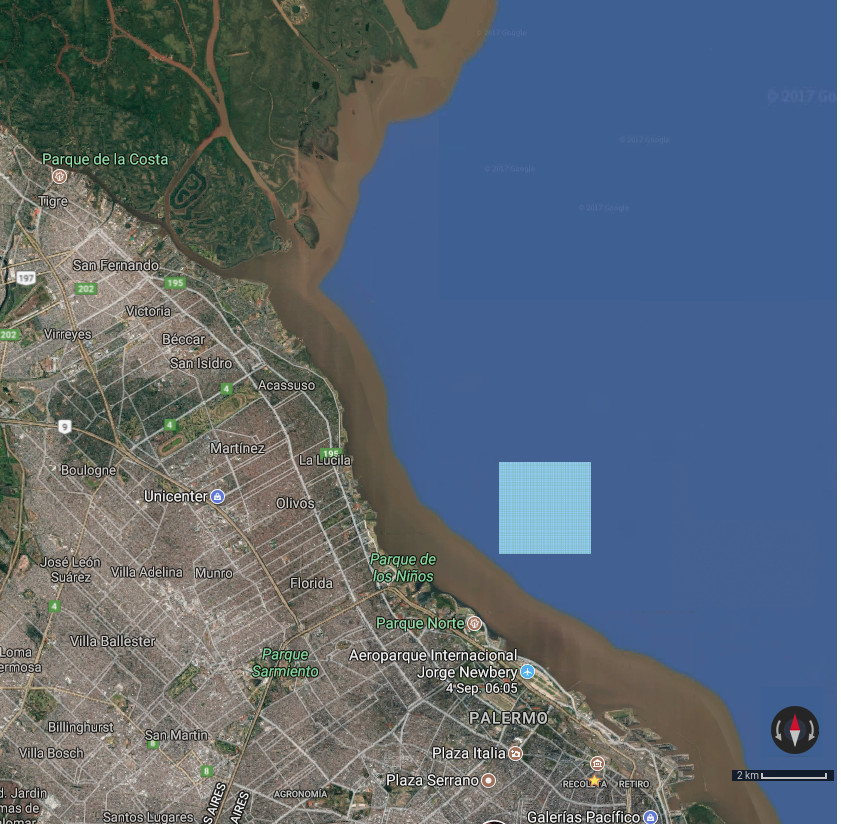}
\caption{Court location in Google Maps \textcopyright\ capture.}
\end{figure}

\subsection{Initial wind state}

Knowing historical weather conditions, it is possible to implement a model to generate initial wind conditions based on an intermediate weather stability range. Fig.~5 shows a representation of the initial state for a a 40000 m${}^2$ court ($4 \times 4$ cells). Arrow length is proportional to the wind speed defined for the cell (we assume the same speed and angle within each cell), and arrow direction represents wind direction.

In the considered location near Buenos Aires, we have that (a) the wind direction is generally E-W, (b) the wind speed tends to be lower at W than at E, and (c) the wind direction tends to be more perpendicular to the coast at W than at E.

We implemented a procedure generating valid initial states according to these premises and certain random variations. The parameters of the procedure are the mean and standard deviation of the incidence angle of winds, and the mean and standard deviation of wind speed. We use Gaussian random number generators for this task.

\begin{figure}[!htb]
\centering
\includegraphics[scale=1.15]{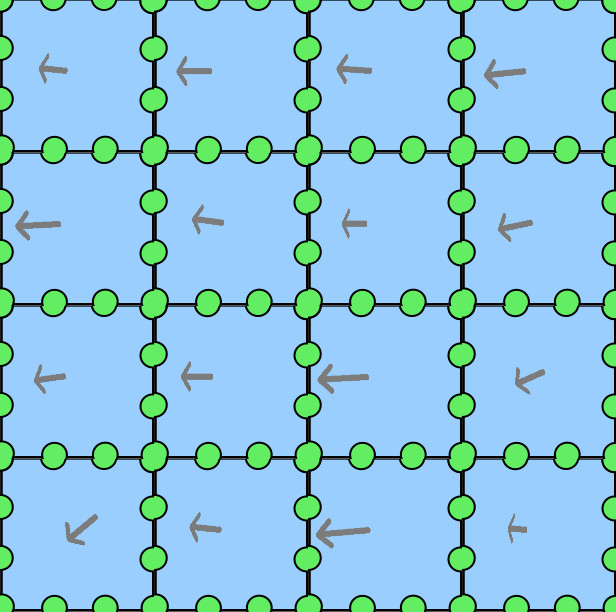}
\caption{Initial state for a $4 \times 4$ court.}
\end{figure}

\subsection{Wind simulation procedure}

Once a initial state $w_0$ is defined by the procedure described in Section~3.1, we perform a simulation process starting from $w_0$. We define a \emph{scenario} to be a time-indexed sequence of winds-state sets for each cell in the court. We take $t_i = 10\:i$ seconds, for $i=0,\dots,n$, to be the time steps for the simulation, hence 360 states are needed in order to simulate an hour of wind states.

The simulation proceeds as follows. We introduce changes into some state in order to produce a new state, by slightly modifying randomly-selected cell states by adding normal Gaussian random numbers (using $\sigma$ as a parameter of the process). This is performed for both wind speed and angle. Once some cells are altered, we implement a \emph{spreading change procedure} to alter neighboring cells until all cells are processed.

After these slight changes are performed, we add \emph{gusts} of winds \cite{3}, a crucial element in regattas. There is no uniform probability of gusts falling all over the court, so we assign probabilities of gust falls to each cell as a new parameter of the court. Gusts of winds have their own parameters: (a) mean time between two consecutive gusts, (b) mean length, (c) standard deviation of the gust length, (d) mean wind angle variation, (e) standard deviation on angle variation, (f) mean wind speed variation, and (g) standard deviation of wind speed variation. If the random variable determines that a gust should appear in the time slot being (parameter (a)), then the cell in which it occurs is randomly selected (according to the gust fall probability for each cell). Gaussian random numbers are generated in order to state the gust particular length (parameters (b) and (c)) and speed and angle variation (remaining parameters). Wind variations are spread using same procedure logic.

\section{Evolutionary algorithm for searching minimum cost paths in simulated scenarios}

Once a set of wind scenarios is produced, we apply a metaheuristic search to find reasonably good solutions. The main objective of this metaheuristic search is to find good paths among all scenarios, so we are not looking for an optimal route in one state neither looking for an optimal route in one scenario (time-dependent solution), but we are looking for robust routes that are good in most scenarios instead.

\subsection{Chromosome design}

We discarded a classical genetic algorithm representation in terms of bit strings since (1) no structure could be found by a genetic workflow in few generations (we tried less than 1000 generations) and (2) no simple crossover operator seems to be possible. Due to this fact, we chose to model individuals (chromosomes) as lists of ordered integer pairs, where each tuple represents a navigation vertex (in terms of Fig.~1). For convenience, this representation excludes the initial and goal vertices. The vertices taken from the lists included within the genes do not include the maneuvering component, in order to let the evolutionary algorithm detect important points where maneuvers should be made.

Each individual does not represent a single route, but a set of routes instead. Within evolutionary algorithms this approach is known as the \emph{Pittsburgh approach} \cite{4}. An individual is modeled with a unique chromosome containing only one non-fixed length gene. Maximum and minimum gene length are parameters of the algorithm that can affect performance. Each locus inside a gene is occupied by a vertex (represented as an ordered pair).

\subsection{Population initialization}

A random initialization schema is not allowed in our case, since feasible individuals are not easily generated and random paths are usually worse than manually-generated routes. Nevertheless, it is important to introduce some level of randomness in the initialization process, so our population initialization procedure includes $s_{w0}$ as an individual and constructs individuals based on paths containing some random vertex. 

\subsection{Individual development and fitness evaluation}

Our framework defines an interface ``Morphogenesis Agent'' whose mission is to transform genotypes into fenotypes. In our case, this amounts to converting lists of ordered pairs into a family of valid paths. This process must generate and evaluate the family of possible valid paths from the individual. For convenience, in this implementation it is useful to evaluate paths and determine individual fitness in the same process in order to take advantage of distributed computing resources.

The fitness function is simple: each path in the family developed by the morphogenesis agent has an associated cost (evaluated during development process). Call $c_{ij}$ to the cost of the path number $j$ within individual $i$. In order to provide a positive fitness value, the fitness function applied to an individual $k$ is defined as $FF(k) = M- \sum_j^{}{\min(c_{kj})}$, with $M \in \mathbb{N}$ an arbitrarily big number.

\subsection{Crossover operators}

Classical one-point or two-point crossover operators were not able to find good paths in our experiments, so we adapted the ideas suggested in \cite{5} to our logic. Our crossover mechanism contemplates two possibilities: (1) parent individuals have a common vertex or (2) not. If they have a common vertex, then a path recombination is made with a random selected \emph{pivot}, chosen among common vertices. Two auxiliary procedures are performed after such a recombination, namely cycle elimination and chromosome reparation. In case (2), a variation of one-point crossover was implemented. Experiments showed that the latter can include some randomness when needed. In our experience, regular to good paths hace vertices in common, so most crossovers apply case (1). Fig.~6 shows an example of our crossover operator combining two parents (I1 and I2) and generating descent (Desc1 and Desc2).

\begin{figure}[!htb]
\centering
\includegraphics[scale=0.42]{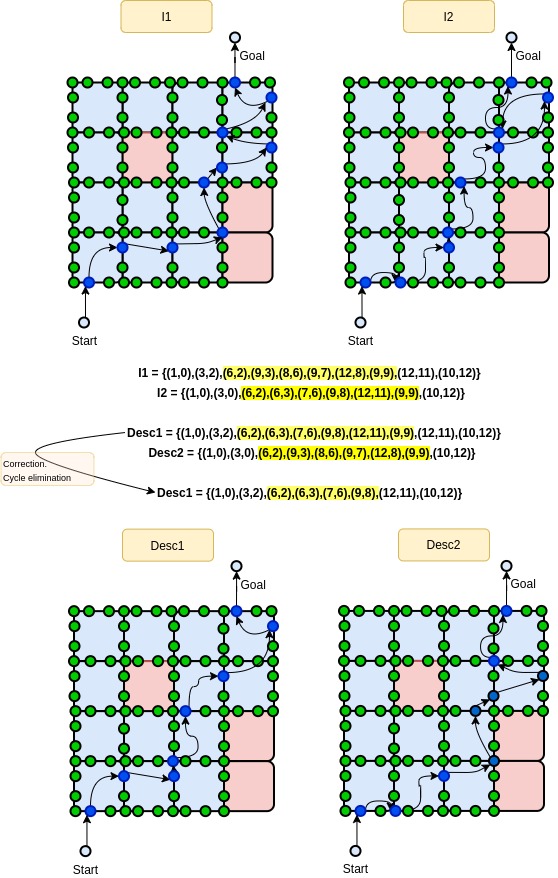}
\caption{Crossover mechanism.}
\end{figure}

\subsection{Mutation operator}

Our mutation operator randomly choses a vertex $v$ of the chromosome, and swaps $v$ with a randomly-chosem vertex $w$. If $v$ is located within the first 20\% of the path, then $w$ is selected close to the starting vertex of the path. If $v$ is located wihtin the last 80\% of the path, then $w$ is chosen closer to the goal vertex of the path.

\section{Evolutionary workflow implementation}

In order to develop an individual and to compute fitness, the evolutionary workflow needs to evaluate every path generated for every individual of the population for every wind scenario. Due to the number of scenarios, this must be processed in a distributed fashion. The rest of the workflow (namely, population initialization, probabilistic roulette selection, crossover execution, mutation execution, and descendant acceptance) is not expensive in terms of computational resources.

Our framework is built on top of Apache Spark framework, is implemented in the Scala language, and it is prepared to run on (private or public) Kubernetes-based clouds. In this particular workflow, we distribute an RDD \cite{6} list with wind scenarios all over the cluster. The process computes a map operation in which the whole population is processed in each executor (each one has a part of the wind scenarios set).

\section{Sizing}

We considered instances with 6.25 km${}^2$ courts. Our cluster has 96 vCPUs and 224GB of total RAM. We planned to assign one instance per vCPU, so we have $w_{spark}=96$ instances. We consider $100$ individuals in the population, $9600$ wind scenarios, and run the evolutionary algorithm for $1000$ iterations. In this case, the cloud will perform $9.6 \times 10^{10}$ path evaluations. With no boosting of any type, a generation takes around 1 hour to be processed in the cluster, keeping 100\% of CPU for most of time.

\section{Results}

Our evaluation process involves taking a holdout set of $H=10$ scenarios not used in the evolutionary algorithm. After the algorithm runs, we take the top 10 solutions found by the evolutionary algorithm ($S_{10}$), and compare $s_{w0}$ with $S_{10}$ in the $H$ scenarios. 

Exact optimal solutions like $s_{w0}$ showed to be good solutions in small court instances. For example, in a 40000 m${}^2$ court, in average, it took 15 generations to find similar solutions to $s_{w0}$. Slightly better solutions were also found. 

In case of larger courts, like a 6.25 km${}^2$ court, $s_{w0}$ turned out not to be a good solution. Every single solution taken from $S_{10}$ is better, and it took about 10 generations to find solutions with similar costs.

The main problem of our implementation is performance. These experiments show that our algorithm can provide robust solutions for real environments, but is expensive in terms of computational resources and running time. This observation motivates the rest of this work.

\section{Boosting performance}

The first approach to improve performance was to maintain a cache. The idea of this cache is to avoid evaluating the same individual twice. With this simple goal, we measure that, in average, a 17\% of individuals evaluation could be avoided and performance was improved accordingly. Given the nature of RDD distribution, the time reduction amounts to 12\% of overall time.

In order to avoid a greater proportion of evaluations we decided to incorporate ML techniques. The assumption is that with enough data, a path may be classified as being ``good'' or ``bad'', so we can only evaluate good paths (hence this is a binary classification problem). Such an approach has a problem associated with evolutionary algorithms, namely that it is important to have a fitness value for each individual. To tackle this issue, we also resort to a mechanism for estimating the cost of a path.

In order to evaluate this approach we experimented with small instances given by medium-sized courts (0.16 km${}^2$), populations of 100 individuals, 200 scenarios, and 100 generations. Based on 10000 evaluations of individuals, we collected 8590 individuals with their costs (14.1\% cached in this run). We also split this set into a train set and keep a hold-out validation set of 20\% of the individuals.

After evaluating alternatives, we decided to implement gradient boosting decision trees (GBDT). In order to keep our framework in Scala-Spark environments, we chose Spark MLLib as the main framework for our ML pipelines. We used LightGBM \cite{7}, one of the most referenced implementations in decision trees which is also compatible with Spark MLLib.

\subsection{Classification algorithm}

In order to determine whether an individual is good or not, we use a threshold around the cost of $s_{w0}$. A solution with cost greater than $s_{w0}+\epsilon$ is considered to be a bad solution.

We transformed inputs into data frames with a binary label (0/1) and a list of features (labeled vertex with ordered natural numbers, from left to right, and bottom up fashion). 

After training and fitting the binary classification GBDT model, we found that the accuracy measurement was always around $0.98$, the precision around $0.99$, the recall around $0.98$, and the sensitivity around $0.91$. After these measurements were confirmed, we performed additional tests in order to rule out the possibility of overfitting.

The confusion matrix in Fig.~7 shows that, in average, we get $165$ true positives and $1518$ true negatives, $14$ false negatives ($0.91\%$ of total ``bad'' solutions) and $16$ false positives ($8.93\%$ of total ``good solutions''). Even when these seem to be very good indicators, false negatives may be of concern. Fortunately, a regression algorithm can recover part of this loss, as the following section shows.

\begin{figure}[!htb]
\centering
\includegraphics[scale=0.6]{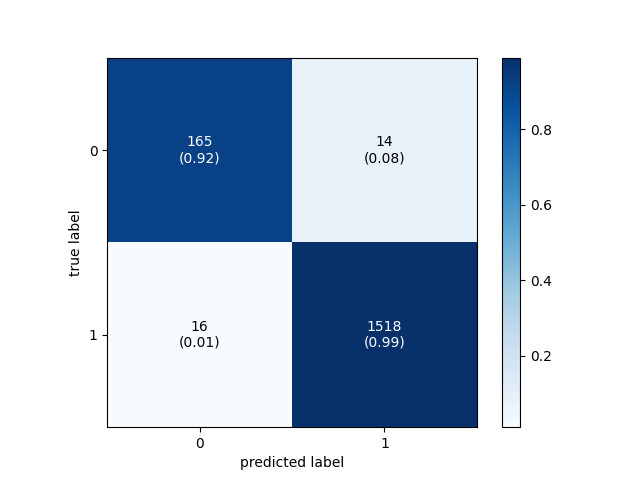}
\caption{Confusion matrix for GDBT binary classification.}
\end{figure}

\subsection{Regression algorithm}

In order to forecast costs of not-so-good solutions, we implemented a regression model with GBDT. The feature engineering was very similar to the description in Section~8.1, but the solution cost (instead of binary values) was used as target.

After training and fitting this model, for a test instance for which we used a threshold of $355$ cost units to divide good solutions from bad solutions, we obtained $MSE=1173$, $RMSE=34.2$, and $MAE=16$. 

An histogram of results shows that $82\%$ (1409 cases) of forecasted (held-out cases) solutions were under $3.5$ cost units. Only $14$ outlier cases ($0.8\%$) were over 150 cost units.

We run several experiments in order to determine how many false negatives were corrected by regression algorithms, and found out that about $95\%$ of the cases are forecasted with less than $50$ cost units of error. 

We analyzed final populations in $100$ generation processes (for a population of $100$ individuals), and found that ``good'' individuals were less than $40$. Taking this proportion into account, in average, our classification algorithm should tag $55$ false negatives for the whole run. Of those $55$ false negatives, $52$ should be recovered with our regression algorithm. So, the actual loss for this run is $3$ ``good'' solutions.

\subsection{In-line version}

After successfully showing that this approach can boost the performance of the evolutionary algorithm, we decided to implement an in-line version in order to train models after each generation is completed. We could observe that accuracy measurement always reached the $0.75$ threshold in 3 or 4 generations, so this suggests that it is possible to use this approach in an in-line way.

For a running instance of 100 generations, this technique allows to reduce the overall running time by $83\%$, with no significant loss of accuracy in terms of solution quality compared to results without ML boosting. Loss of accuracy was measured to be in the range of $0.01-0.05\%$ of solution cost.

\subsection{Tuning of hyperparameters}

Although decision trees and gradient boosting have a good tunability level \cite{8}, we found that default values worked very well with a good number of individuals (8590). There was no practical difference between default and tuned hyperparameter values for the test proposed in this section, with the exception of the in-line version (Section~8.3), in which the data set can be poorly populated in the very early generations. In this case we apply a cross-validation with $k$-folds ($k=5$) pipeline in order to tune the hyperparameters. A simple grid search approach was enough to tune the hyperparameters in early steps of in-line executions, running the tuning process after each generation. Special care was taken in order to avoid overfitting at very early stages. 

The search space was reduced by removing hyperparameters for which tunability showed to be poor (considering the in-line running). As a result, hyperparameters for our GBDT are the number of leaves, the minimal number of data in one leaf, $\lambda_1$, and $\lambda_2$ \cite{9}.

\section{Conclusions}

In this work we have presented an application of an evolutionary algorithm for finding robust solutions, with a distributed component in order to evaluate solutions in a large number of scenarios. The implementation of such a procedure relies on a distributed framework, and allows to find competitive and robust solutions for real-sized instances. We have also described the incorporation of ML techniques to the basic implementation, in order to boost performance by not evaluating all individuals. Our experiments show that this can be achieved with a reasonable effort and a small impact on the obtained solutions. It would be interesting to explore whether these techniques can be applied in other big data optimization environments.

\bibliographystyle{named}

\end{document}